\documentclass[10pt,twocolumn,letterpaper]{article}

\usepackage{multirow}
\usepackage{tabularx}
\usepackage{float}
\usepackage{stfloats}
\usepackage[section]{placeins}
\usepackage[pagenumbers]{cvpr} % To force page numbers, e.g. for an arXiv version

\usepackage{microtype}
\usepackage{graphicx}
\usepackage{subcaption}
\usepackage{booktabs} % for professional tables

\usepackage{xcolor}
\usepackage{color}
\usepackage[table]{xcolor}
\usepackage{multirow}
\usepackage{tabularx}
\usepackage{float}
\usepackage{algorithm}
\usepackage{algpseudocode}
\usepackage{setspace}
\usepackage{bm}
\usepackage[pagebackref=true,breaklinks=true,letterpaper=true,colorlinks,bookmarks=false]{hyperref}

\usepackage{booktabs}
\usepackage{multirow}
\usepackage{amssymb}

\definecolor{cvprblue}{rgb}{0.21,0.49,0.74}
\definecolor{lightpurple}{rgb}{0.902, 0.863, 0.922}
\definecolor{lightgreen}{rgb}{0.863, 0.902, 0.863}
\definecolor{lightpink}{rgb}{0.980, 0.941, 0.941}

\newcommand{\best}[1]{\textbf{\textcolor{red}{#1}}}
\newcommand{\sbest}[1]{\textbf{\textcolor[HTML]{00B0F0}{#1}}}

\def\confName{CVPR}
\def\confYear{2026}

\title{DriftSR: One-Step Real-World Image Super-Resolution via Distribution Drifting}
\author{
Wei Zhu$^{1,3}$ \quad
Kai Zhang$^{2,}\thanks{Corresponding authors.}$ \quad
Yu Zheng$^{1}$ \quad
Zhaopeng Yang$^{3}$ \quad
Lei Luo$^{1}$ \quad
Yong Guo$^{4}$ \quad
Jian Yang$^{1,2,*}$ \\
$^{1}$Nanjing University of Science and Technology \qquad
$^{2}$Nanjing University \qquad 
$^{3}$Huawei \\
$^{4}$South China University of Technology \qquad
}

\begin{document}

\setlength{\abovedisplayskip}{3pt}
\setlength{\belowdisplayskip}{2pt}
\maketitle

\begin{abstract}
One-step real-world image super-resolution (Real-ISR) offers efficient inference, but recovering realistic and perceptually rich details often relies on score distillation or adversarial learning, introducing additional trainable components and making optimization more cumbersome. To this end, we propose DriftSR, a one-step Real-ISR framework that leverages pretrained diffusion priors through distribution drifting. Specifically, we perform drifting in the frozen intermediate representation space of a pretrained diffusion model, without introducing an additional task-specific feature encoder. Building on this space, we introduce Spatial Feature Drifting, which treats spatial features rather than entire images as distributional samples, enabling denser supervision for distribution alignment. To mitigate structural deviations, we further introduce Structure-Modulated Guidance, which adaptively refines drifting guidance according to local structural consistency with the LQ input. Consequently, DriftSR optimizes only the one-step generator, without auxiliary distillation branches or adversarial discriminators. Extensive experiments on three real-world benchmarks demonstrate that DriftSR delivers high-quality super-resolution reconstruction with efficient one-step inference.
\end{abstract}    
\section{Introduction}
\label{sec:introduction}

Real-world image super-resolution (Real-ISR)~\cite{2021Designing, 2021Real,2018Learning, deep_unfolding,DVAR} aims to recover high-quality (HQ) images from low-quality (LQ) observations affected by unknown and complex degradations. The mapping from LQ observations to HQ images is inherently ambiguous, and pixel-level reconstruction objectives often lead to over-smoothed results~\cite{2017Photo,2017EnhanceNet}. Therefore, recovering natural image details while maintaining content and structural fidelity to the input remains a central challenge in Real-ISR~\cite{2024OSEDiff, 2024stablesr}.

Recently, large-scale pretrained diffusion models have significantly advanced Real-ISR by providing strong generative priors for natural images~\cite{2024stablesr,supir,wu2024seesr,dong2024tsd}. As the field shifts from iterative diffusion-based restoration toward efficient one-step generation, a key challenge is to recover realistic and perceptually rich details within a single forward pass. To address this challenge, existing diffusion-based one-step SR methods~\cite{2024OSEDiff,wang2026strsr,s3diff,li_d3SR} often employ score distillation or adversarial learning to enhance perceptual details. Score distillation approaches~\cite{vsd,2024OSEDiff,dong2024tsd} typically rely on an additional teacher model and trainable LoRA branches, with the LoRA parameters optimized under diffusion-based supervision throughout training. Adversarial approaches~\cite{gan,s3diff,li_d3SR} instead introduce an additional discriminator and may suffer from unstable optimization. Despite their effectiveness in improving perceptual quality, both paradigms require auxiliary trainable components and additional objectives, increasing the complexity of the training pipeline. This motivates a simpler alternative that optimizes only the one-step generator while effectively leveraging generative priors.

The recently proposed Drifting Model~\cite{drifting_model} introduces an alternative to score distillation and adversarial learning for distributional generative optimization. It models the relation between source and target distributions in a pretrained feature space and constructs a drifting field guiding generated samples toward the target distribution. By operating on sample distributions without an additional discriminator or a separately optimized distillation branch, distribution drifting presents a potential alternative for one-step Real-ISR.

Nevertheless, directly extending distribution drifting to one-step Real-ISR is non-trivial. \textbf{\textit{First}}, the effectiveness of the drifting field depends strongly on the feature space used to measure sample relationships, while obtaining a suitable representation may require additional design for Real-ISR. \textbf{\textit{Second}}, drifting at the image level constructs negative samples by excluding the current sample and therefore requires additional samples under the same condition. Such a requirement can be naturally satisfied in image generation conditioned on class labels or text prompts, where multiple images can share the same condition. In contrast, Real-ISR provides only one paired HQ target for each LQ input, leaving no additional sample for negative construction and thus preventing direct application of drifting at the image level. \textbf{\textit{Third}}, Real-ISR requires both high perceptual quality and fidelity to the LQ input. Excessive drifting may encourage visually plausible textures that are inconsistent with the input, resulting in incorrect details and structural distortions.

\begin{figure*}[t]
    \centering
    \includegraphics[width=\linewidth]{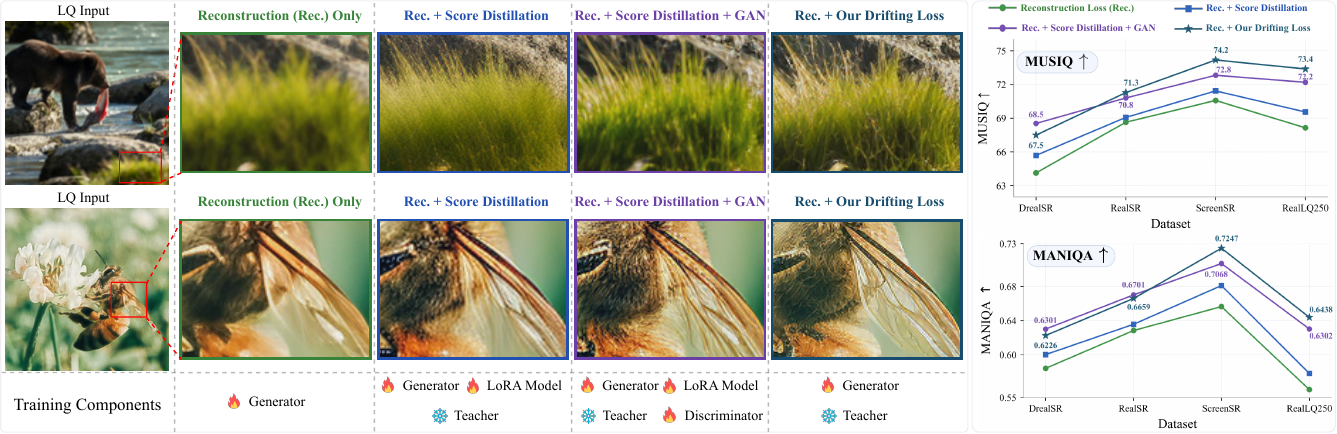}
    % \vspace{-9pt}
    \caption{Qualitative and quantitative comparison of different training objectives. \textbf{Left}: Reconstruction-only training with MSE and LPIPS produces over-smoothed results. Adding variational score distillation~\cite{vsd} improves perceptual details, while further incorporating the GAN objective from OMGSR~\cite{omgsr} enhances texture realism at the cost of additional trainable components. Our drifting loss achieves comparable or better perceptual quality without introducing additional trainable modules. \textbf{Right}: MANIQA and MUSIQ scores on four real-world datasets, where drifting remains competitive with joint score distillation and adversarial training.}
    \label{fig:teaser}
    \vspace{-4mm}
\end{figure*}

To address these challenges, we propose \textbf{DriftSR}, a one-step Real-ISR framework that adapts distribution drifting for perceptual detail enhancement. Inspired by Teacher-Feature Drifting (TFD)~\cite{teacher_drifting}, we employ the pretrained diffusion model as a frozen feature extractor and construct the drifting space using its intermediate representations. This design eliminates the need for an additional task-specific feature network while providing a representation space naturally aligned with the diffusion prior used by the generator. We further introduce \textbf{Spatial Feature Drifting}, which reformulates distribution modeling from the image level to the spatial feature level. By treating each spatial location as an individual sample, this formulation enables negative sample construction within each image and provides denser observations for modeling local texture distributions. To balance perceptual enhancement and input fidelity, we propose \textbf{Structure-Modulated Guidance}. Based on local structural consistency with the LQ input, this guidance adaptively modulates drifting supervision across spatial locations, thereby suppressing visually plausible but structurally inconsistent details. As shown in Figure~\ref{fig:teaser}, these designs collectively enable DriftSR to leverage the pretrained diffusion prior for perceptual detail enhancement while preserving structural fidelity, with optimization restricted to the one-step generator.

Our contributions are summarized as follows:

\begin{itemize}
  \item We propose DriftSR, a one-step Real-ISR framework that introduces Spatial Feature Drifting to model distributions over spatial features and enhance perceptual details without additional trainable components.
  
  \item We introduce Structure-Modulated Guidance to adaptively modulate drifting supervision based on local structural consistency with the LQ input, suppressing structural inconsistencies while preserving perceptually rich textures.
  
  \item Extensive experiments on real-world benchmarks demonstrate that our DriftSR achieves superior perceptual quality while maintaining efficient one-step inference.
\end{itemize}

\section{Related work}
\label{sec:related_work}

\noindent
\textbf{One-Step Diffusion-based Real-ISR.} To reduce diffusion model inference costs in Real-ISR, recent studies have increasingly explored one-step generative models replacing iterative generation with a single forward pass~\cite{resshift,CTMSR,2024OSEDiff}. These methods aim to substantially improve inference efficiency while preserving high perceptual restoration quality. Existing one-step Real-ISR methods~\cite{mfsr,CTMSR,dong2024tsd} commonly adopt distribution or trajectory alignment objectives during training. Among them, distribution alignment is often achieved through score distillation, which constructs supervision from the score discrepancy between a frozen diffusion teacher and a trainable LoRA branch. Some methods further incorporate adversarial learning~\cite{s3diff,li_d3SR,smfsr} to enhance perceptual quality, introducing an additional discriminator. Although these strategies achieve strong restoration performance, the auxiliary trainable components and additional objectives increase the complexity of the training pipeline. Therefore, we explore a simpler alternative that leverages pretrained diffusion priors to guide one-step Real-ISR, without auxiliary distillation branches or discriminators.

\noindent
\textbf{Drifting Model.} Drifting Model~\cite{drifting_model} introduces a new paradigm for one-step generation by moving distribution transport from inference to training. Unlike diffusion or flow-based models~\cite{sd21-base,sd3,FLUX.2-klein-4B} that rely on iterative inference, the drifting model learns a one-step generator whose output distribution is progressively updated during training under a drifting field defined by a kernel. The drifting field attracts generated samples toward nearby target samples while repelling them from the current model distribution, thereby progressively aligning the two distributions. Since distribution transport is performed during training, inference requires only a single forward pass. The effectiveness of the drifting model depends on the representation space used to measure sample relationships~\cite{drifting_model}. For high-dimensional images, distances measured in pixel or latent space may not adequately capture perceptual and semantic similarity. The drifting model therefore uses pretrained representations to construct informative feature spaces, making the choice of feature space important for effective distribution alignment.

% 漂移模型为单步生成提供了一种新的建模范式，其核心思想是将分布传输过程由推理阶段转移至训练阶段。不同于扩散模型或流模型中依赖迭代求解逆向 SDE/ODE 的生成方式，漂移模型直接学习单步生成器，并在训练过程中利用核函数定义的漂移场持续更新生成分布。具体而言，该漂移场通过将生成样本吸引至邻近的目标样本，同时施加来自当前模型分布的排斥作用，引导生成分布逐步逼近目标数据分布。由于分布演化主要发生在训练阶段，模型在测试时无需执行显式的迭代传输过程，从而天然支持单次前向生成。

% 漂移的有效性很大程度上取决于用于度量样本关系的表示空间。对于高维图像，像素空间或简单潜在空间中的距离往往难以刻画感知和语义相似性。因此，现有漂移方法通常借助预训练表示模型构建更具判别性的特征空间，并在其中进行分布建模。如何根据具体任务选择或构建合适的漂移特征空间，是影响漂移监督有效性的关键因素之一。。TFD 直接在用于初始化生成器的扩散教师特征空间中计算漂移损失，从而避免额外的特征空间设计。本文采用相同的教师空间，并针对 Real-ISR 重新设计分布建模粒度与监督机制，以更好地建模局部纹理并减少结构偏移。

% 漂移方向的有效性在很大程度上取决于用于度量样本关系的表示空间。对于高维图像，直接在像素空间或简单潜在空间中计算样本距离，往往难以充分刻画感知层面与语义层面的相似性。因此，现有漂移方法通常借助预训练表示模型构建更具判别性的特征空间，并在该空间中估计样本之间的分布关系。由此可见，如何根据具体任务选择或构建合适的漂移特征空间，是影响漂移监督有效性的关键因素之一。TFD 将 drift loss 直接构建在与生成器初始化一致的扩散教师特征空间中，从而避免了额外的特征空间设计。本文同样采用该教师空间，但进一步针对 Real-ISR 的特点，对 drifting 的分布建模粒度与监督机制进行重新设计，以增强局部纹理建模并抑制结构偏移。

\section{Method}
\label{sec:method}

\subsection{Preliminary}
\label{sec:preliminary}

\noindent\textbf{Drifting Model.} Drifting Model formulates generative modeling as the gradual evolution of the model distribution during training. Let $f_\theta$ denote a one-step generator that maps a noise variable $\epsilon \sim p_\epsilon$ to a generated sample:
\begin{equation}
    x = f_\theta(\epsilon), \qquad x \sim q_\theta,
\end{equation}
where $q_\theta = (f_\theta)_\# p_\epsilon$ denotes the pushforward distribution induced by $f_\theta$. Unlike diffusion or flow-based models, which transport samples along an iterative trajectory at inference time, drifting models progressively update $q_\theta$ by optimizing the generator during training. Consequently, the learned generator supports one-step generation through a single forward pass. To guide the evolution of the generated distribution, Drifting Model defines a vector field determined by the target distribution $p$ and the current model distribution $q$:
\begin{equation}
    V_{p,q}(x)
    =
    V_p^+(x) - V_q^-(x).
    \label{eq:drift_field}
\end{equation}
The attraction component is defined as
\begin{equation}
    V_p^+(x)
    =
    \frac{
    \mathbb{E}_{y^+ \sim p}
    \left[
    k(x,y^+) (y^+ - x)
    \right]
    }{
    \mathbb{E}_{y^+ \sim p}
    \left[
    k(x,y^+)
    \right]
    },
\end{equation}
whereas the repulsion component is defined as
\begin{equation}
    V_q^-(x)
    =
    \frac{
    \mathbb{E}_{y^- \sim q}
    \left[
    k(x,y^-) (y^- - x)
    \right]
    }{
    \mathbb{E}_{y^- \sim q}
    \left[
    k(x,y^-)
    \right]
    },
\end{equation}
where $k(\cdot,\cdot)$ is a similarity kernel, typically defined using Laplace distance logits of the form  $-\lVert y-x\rVert/\tau$, with $\tau$ denoting a temperature parameter. The normalized kernel weighting makes $V_p^+(x)$ and $V_q^-(x)$ local mean-shift directions determined by nearby real and generated samples, respectively. Their difference therefore provides a drifting direction that moves the current model distribution toward the target distribution. When $p=q$, the two local mean-shift fields coincide, causing the drifting field to vanish and yielding the equilibrium condition:
\begin{equation}
    p=q
    \quad \Rightarrow \quad
    V_{p,q}(x)=0.
\end{equation}

During training, the drifting field is used to construct a transported target for each generated sample $x=f_\theta(\epsilon)$:
\begin{equation}
    \tilde{x}
    =
    x + V_{p,q_\theta}(x).
\end{equation}
The generator is then optimized to regress toward this target using a stop-gradient operation:
\begin{equation}
    \mathcal{L}_{\rm drift}
    =
    \mathbb{E}_{\epsilon \sim p_\epsilon}
    \left[
    \left\|
    f_\theta(\epsilon)
    -
    \operatorname{sg}
    \left(
    f_\theta(\epsilon)
    +
    V_{p,q_\theta}
    \left(
    f_\theta(\epsilon)
    \right)
    \right)
    \right\|^2
    \right],
    \label{eq:drift}
\end{equation}

The quality of the drifting field depends strongly on the representation used to measure sample similarity. For high-dimensional images, distances in pixel or latent space may not adequately capture semantic similarity. Therefore, the drifting field is typically constructed in an informative feature space provided by a pretrained representation model.

\noindent\textbf{Problem Formulation.} 
The Real-ISR task aims to reconstruct a HQ image $x_{H}$ from a LQ input $x_{L}$ that suffers from complex and unknown degradations. Given a parameterized restoration model $G_{\theta}$, Real-ISR can be formulated as the following optimization problem:
\begin{equation}
\min_{\theta}
\left(
\mathcal{L}_{\mathrm{rec}}(G_{\theta}(x_{L}), x_{H}),\;
\mathcal{L}_{\mathrm{reg}}(G_{\theta}(x_{L})
\right),
\label{eq:sr_objective}
\end{equation}
where $\mathcal{L}_{\mathrm{rec}}$ denotes the reconstruction term, enforcing consistency between the prediction $\hat{x}_{H} = G_{\theta}(x_{L})$ and ground truth $x_{H}$ using $L_2$ or LPIPS~\cite{lpips}. The term $\mathcal{L}_{\mathrm{reg}}$ incorporates natural image priors by guiding the generated distribution toward the real image distribution. Common choices include Variational Score Distillation (VSD)~\cite{vsd} and GAN-based objectives~\cite{wgan}. However, VSD typically requires a trainable score model and a pretrained diffusion teacher, whereas GAN-based regularization introduces a discriminator and may cause unstable adversarial training. We instead adopt the drifting objective as the sole distribution regularizer, avoiding these trainable components while guiding the generated distribution toward the real image distribution.

\subsection{Overview of DriftSR}
\label{sec:overview}
% As illustrated in Figure~\ref{fig:method}, given an LQ image $x_L$, the generator $G_\theta$ predicts the restored image $\hat{x}_H$ and is supervised by the reconstruction loss $\mathcal{L}_{\mathrm{rec}}$, with the paired HQ image $x_H$ serving as the target. A frozen pretrained teacher extracts multilevel latent features $F$ and $\hat{F}$ from $x_H$ and $\hat{x}_H$, respectively. Their spatial features serve as positive samples $P^\ell$ and negative samples $Q^\ell$ to construct the drifting field $V^\ell$ at each selected teacher layer (Sec.~\ref{subsec:drifting_loss}). The same procedure is applied to the DWT low-frequency components $x_H^{\mathrm{low}}$ and $\hat{x}_H^{\mathrm{low}}$ to obtain the low-frequency drifting field $V_{\mathrm{low}}^\ell$. A structure modulation map $\mathbf{M}$ adaptively combines the two drifting fields into $V_{\mathrm{SMG}}^\ell$, which is used to construct the layer-wise drifting objective. The final drifting loss $\mathcal{L}_{\mathrm{drift}}$ is obtained by aggregating the objectives across all selected teacher layers (Sec.~\ref{subsec:smg_guidance}). During training, $G_\theta$ is jointly optimized with $\mathcal{L}_{\mathrm{rec}}$ and $\mathcal{L}_{\mathrm{drift}}$.

\begin{figure*}[t]
    \centering
    \includegraphics[width=\linewidth]{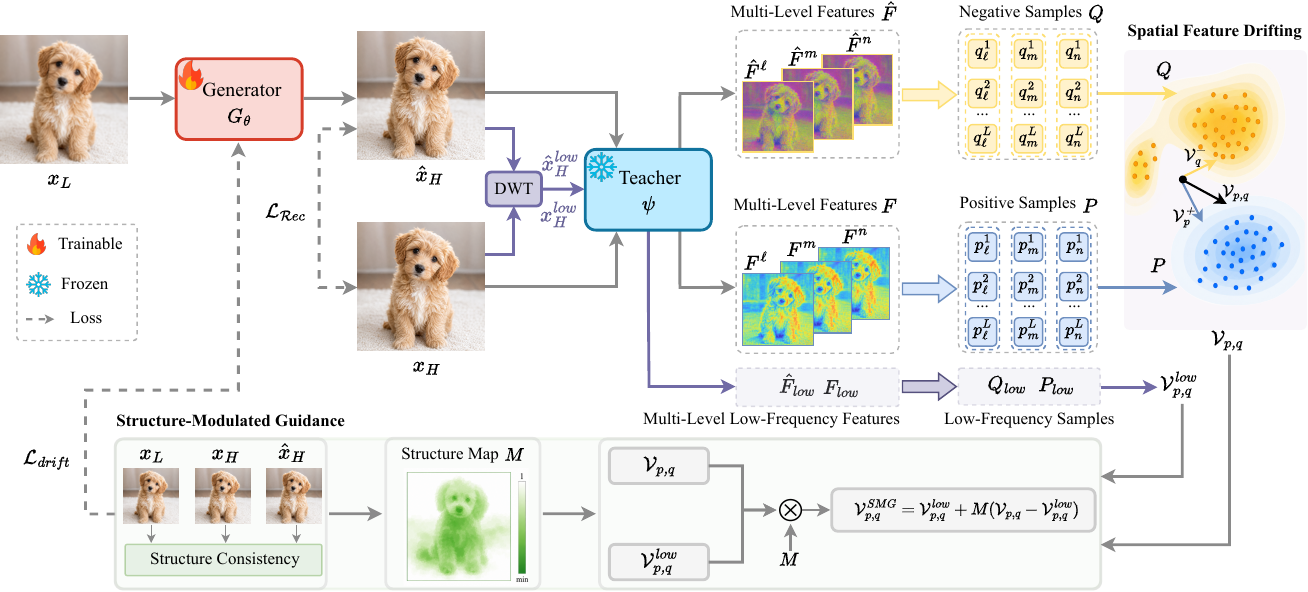}
    % \vspace{-9pt}
    \caption{Overview of DriftSR training pipeline. Given an LQ image $x_L$, $G_\theta$ produces $\hat{x}_H$ under reconstruction supervision from the corresponding HQ image $x_H$. A frozen pretrained teacher extracts multi-level latent features from $x_H$ and $\hat{x}_H$. These features are converted into positive and negative spatial samples to construct the original drifting field. The same procedure is applied to their DWT low-frequency components to construct a low-frequency drifting field. A structure map $M$ adaptively combines the two drifting fields to obtain a structure modulated drifting field for computing the drifting loss. During training, only $G_\theta$ is optimized using the reconstruction and drifting losses.}
    \label{fig:method}
    \vspace{-2mm}
\end{figure*}

As illustrated in Figure~\ref{fig:method}, given an LQ image $x_L$, the generator $G_\theta$ predicts the restored image $\hat{x}_H$ and is supervised by the reconstruction loss $\mathcal{L}_{\mathrm{rec}}$, with the paired HQ image $x_H$ serving as the target. A frozen pretrained teacher extracts multilevel latent features $F$ and $\hat{F}$ from $x_H$ and $\hat{x}_H$, respectively. At each selected layer, the corresponding spatial features serve as positive samples $P$ and negative samples $Q$ to construct the drifting field $V_{p,q}$ (Sec.~\ref{subsec:drifting_loss}). The same procedure is applied to the DWT~\cite{dwt_theory} low-frequency components $x_H^{\mathrm{low}}$ and $\hat{x}_H^{\mathrm{low}}$ to construct the low-frequency drifting field $V_{p,q}^{\mathrm{low}}$. A structure modulation map $\mathbf{M}$ then adaptively combines the two drifting fields to obtain $V_{p,q}^{\mathrm{SMG}}$, which is used to construct the layer-wise drifting objective. The final drifting loss $\mathcal{L}_{\mathrm{drift}}$ is obtained by aggregating the objectives across all selected teacher layers (Sec.~\ref{subsec:smg_guidance}). During training, $G_\theta$ is optimized using both $\mathcal{L}_{\mathrm{rec}}$ and $\mathcal{L}_{\mathrm{drift}}$.

\subsection{Spatial Feature Drifting}
\label{subsec:drifting_loss}
The effectiveness of distribution drifting depends strongly on the feature space used to measure sample similarity. Inspired by TFD \cite{teacher_drifting}, we employ a frozen pretrained diffusion model as the feature extractor and construct the drifting field in its intermediate feature space. This design offers two advantages. First, it avoids the need for a handcrafted or separately trained feature extractor specifically tailored to Real-ISR. Second, the generator is initialized from the same pretrained diffusion model and therefore inherits a compatible representation hierarchy. This shared initialization naturally aligns the drifting feature space with the representations inherited by the generator, enabling more consistent measurement of sample similarity during optimization.

Concretely, for the reconstructed image $\hat{x}_H$ and its HQ image $x_H$, where $\hat{x}_H, x_H \in \mathbb{R}^{C \times H \times W}$, let $\psi$ denote the parameters of the frozen teacher, and let $E_{\psi}$ and $F_{\psi}$ denote its VAE encoder and denoising network, respectively. Both images are first encoded into the latent space:
\begin{equation}
    \hat{z}=E_{\psi}(\hat{x}_H),
    \qquad
    z=E_{\psi}(x_H).
\end{equation}

The resulting latent representations are then perturbed using the same Gaussian noise at noise strength $t$:
\begin{equation}
    \hat{z}_{t}
    =
    \hat{z}+t\epsilon,
    \quad
    z_{t}
    =
    z+t\epsilon,
    \quad
    \epsilon
    \sim\mathcal{N}(0,I).
\end{equation}

The perturbed latents are subsequently fed into the frozen denoising network.
Let $h_{F_{\psi}}^{\ell}(\cdot)$ denote the intermediate representation extracted from selected layer $\ell$. The corresponding teacher features are extracted as
\begin{equation}
    \hat{F}^{\ell}
    =
    h_{F_{\psi}}^{\ell}(\hat{z}_{t},t,c),
    \qquad
    F^{\ell}
    =
    h_{F_{\psi}}^{\ell}(z_{t},t,c),
\end{equation}
where
$\hat{F}^{\ell},F^{\ell}
\in\mathbb{R}^{C_{\ell}\times H_{\ell}\times W_{\ell}}$, and $c$ denotes the textual information extracted from the input image.

Direct image-level drifting treats the representation of each image as a single sample, which limits its applicability to Real-ISR because each LQ input is associated with only one paired HQ reference during training. To overcome this limitation, we exploit the spatial structure preserved in the teacher features and treat spatial feature vectors as individual drifting samples. Specifically, at each selected layer $\ell$, we unfold the spatial dimensions of the feature maps such that each spatial feature vector forms an individual sample:
\begin{equation}
    Q^{\ell}
    \in
    \mathbb{R}^{N_{\ell}\times C_{\ell}},
    \qquad
    P^{\ell}
    \in
    \mathbb{R}^{N_{\ell}\times C_{\ell}},
\end{equation}
where $N_{\ell}=H_{\ell}W_{\ell}$ denotes the number of spatial samples, and $Q^{\ell}$ and $P^{\ell}$ are constructed from the spatial features $\hat{F}^{\ell}$ and $F^{\ell}$, respectively. This spatial formulation converts each training pair into $N_{\ell}$ feature samples, allowing positive and negative sample sets to be constructed within a single pair. It also provides denser observations for distribution modeling and captures local feature variations that are important for perceptual restoration.

Following the drifting formulation introduced in the preliminaries, $P^{\ell}$ and $Q^{\ell}$ jointly define the drifting field acting on $Q^{\ell}$ in the feature space of layer $\ell$:
\begin{equation}
    V^{\ell}
    =
    V_{P^{\ell},Q^{\ell}}
    \left(
    Q^{\ell}
    \right).
\end{equation}

By modeling distributions over spatial features rather than entire image representations, this formulation enables distribution drifting within each paired training example despite the limited number of image-level references.

\subsection{Structure-Modulated Guidance}
\label{subsec:smg_guidance}

\begin{table*}[ht!]
% \fontsize{12pt}{14.8pt}\selectfont
\centering
\vspace{-2mm}
\caption{A comprehensive comparison with state-of-the-art methods on three real-world datasets. For VOSR, we report the one-step variant with 1.4B parameters. The best and second-best results under each metric are marked in \textcolor{red}{\textbf{red}} and {\color[HTML]{00B0F0}\textbf{blue}} respectively.}
\fontsize{7pt}{8.4pt}\selectfont
\setlength{\tabcolsep}{3.2pt}
\renewcommand{\arraystretch}{1.13}

\resizebox{\textwidth}{!}{
\begin{tabular}{@{}c | c | c c  c c | c c c c c c c c
% >{\columncolor{lightpink}}c}
>{\columncolor{gray!10}}c}
% \toprule
% \textbf{Datasets} &
% \textbf{Metrics} &
% \textbf{DiT4SR} &
% \textbf{StableSR} &
% \textbf{SeeSR} &
% \textbf{ResShift} &
% \textbf{CTMSR} &
% \textbf{OSEDiff} &
% \textbf{PiSA-SR} &
% \textbf{TSD-SR} &
% \textbf{HYPIR} &
% \textbf{VOSR} &
% \textbf{FiDeSR} &
% \textbf{StrSR} &
% \textbf{DriftSR} \\
% \hline
\toprule

\textbf{Datasets} &
\textbf{Metrics} &
\textbf{DiT4SR} &
\textbf{StableSR} &
\textbf{SeeSR} &
\textbf{ResShift} &
\textbf{CTMSR} &
\textbf{OSEDiff} &
\textbf{PiSA-SR} &
\textbf{TSD-SR} &
\textbf{HYPIR} &
\textbf{VOSR} &
\textbf{FiDeSR} &
\textbf{StrSR} &
\textbf{DriftSR}
\\

\multicolumn{2}{c|}{\textbf{\textit{T2I Prior}}}
&
\scriptsize \textbf{\textit{SD3.5}} &
\scriptsize \textbf{\textit{SD2}} &
\scriptsize \textbf{\textit{SD2}} &
\scriptsize \textbf{\textit{$-$}} &
\scriptsize \textbf{\textit{$-$}} &
\scriptsize \textbf{\textit{SD2.1}} &
\scriptsize \textbf{\textit{SD2.1}} &
\scriptsize \textbf{\textit{SD3.5}} &
\scriptsize \textbf{\textit{SD2.1}} &
\scriptsize \textbf{\textit{$-$}} &
\scriptsize \textbf{\textit{SD2.1}} &
\scriptsize \textbf{\textit{FLUX}} &
\scriptsize \textbf{\textit{SD2.1}} \\

\hline

\multirow{9}{*}{\begin{tabular}[c]{@{}c@{}}
\textit{\textbf{DRealSR}}\end{tabular}}

& PSNR $\uparrow$ & 25.7966 &28.030  &28.073 &\sbest{28.693} &28.682 &27.915 &28.318 &27.769 &25.932 & 27.824 & \best{29.034} &25.0242& 27.393\\
& SSIM $\uparrow$ & 0.6822 &0.7536  &0.7684 &\sbest{0.7875} &0.7839 &0.7833 & 0.7804 &0.7559 &0.7197 & 0.7427 & \best{0.7942} &0.7297& 0.7584 \\
& LPIPS $\downarrow$ & 0.3677 &0.3284 &0.3173 &0.3525 &0.3236 &0.2968 &\sbest{0.2960} &0.2967 &0.3371 & 0.3259 & \best{0.2811} &0.3052& 0.3379\\
& CLIPIQA $\uparrow$ & 0.6624 &0.6356 &0.6909 &0.5404 &0.6517 &0.6974 &0.6971  &\sbest{0.7344} &0.6392 & 0.5914  & 0.6927 &0.6435& \best{0.7442}\\
& MUSIQ $\uparrow$ & 64.3893&58.512 &65.080 &52.373 &59.766 &64.691 &66.113 &\sbest{66.599} &61.069 & 66.025  & 65.359 &64.086& \best{67.490} \\
& MANIQA $\uparrow$ & \sbest{0.6218} &0.5594 &0.6051 &0.4760 &0.4853 &0.5903 &0.6160 &0.5874 &0.6053  & 0.6052  & 0.6184 &0.6129& \best{0.6226}\\
& TOPIQ $\uparrow$ & 0.5758 &0.5323 &\sbest{0.6575} &0.4669 &0.5960 &0.6002 &0.6333 &0.6182 &0.5742 & 0.6022 & 0.6043 &0.5782& \best{0.6970}\\
& QALIGN $\uparrow$ & 3.3353 &3.0614 &3.5882 &2.8515 &3.0299 &3.5450 &3.5838 &\sbest{3.6041} &3.5204 & 3.4570 & 3.6243 &3.5902& \best{3.6391} \\

\hline

% =====RealSR=====z
\multirow{9}{*}{\begin{tabular}[c]{@{}c@{}}
\textit{\textbf{RealSR}}\end{tabular}}

& PSNR $\uparrow$ & 23.4421 &24.645 &25.149 &\best{26.382} &25.992 &25.148 &25.503 &24.806 &22.826 & 25.229  & \sbest{26.159} &22.927& 24.357 \\
& SSIM $\uparrow$ & 0.6647 &0.7080 &0.7211 &0.7347 &\best{0.7546} &0.7338 &0.7418 &0.7172 &0.6783 & 0.7175 & \sbest{0.7484} &0.7065& 0.7016 \\
& LPIPS $\downarrow$ & 0.3225 &0.3002 &0.3007 &0.3159 &0.2896 &0.2920 &\sbest{0.2672} &0.2743 &0.3087 & 0.2732 & \best{0.2611} &0.2670& 0.3287 \\
& CLIPIQA $\uparrow$ & 0.6404 &0.6234 &0.6699 &0.5505 &0.6355 &0.6687 &0.6699 &\sbest{0.7162} &0.6390 & 0.5625 & 0.6821 &0.6430& \best{0.7334}\\
& MUSIQ $\uparrow$ & 68.4939 &65.883 &69.819 &60.225 &64.415 &69.087 &70.147 &\sbest{71.179} &66.258  & 70.584 & 69.579 &68.501& \best{71.283}\\
& MANIQA $\uparrow$ & \sbest{0.6635} &0.6230 &0.6450 &0.5402 &0.5268 &0.6337 &0.6551 &0.6347 &0.6436 & 0.6443 & 0.6631 &0.6502& \best{0.6659}\\
& TOPIQ $\uparrow$ & 0.6057 &0.5748  &\sbest{0.6890} &0.5191 &0.5998 &0.6251 &0.6374 &0.6580 &0.6069 &  0.6392 & 0.6142 &0.6228& \best{0.7275} \\
& QALIGN $\uparrow$ & 3.4216 &3.2767 &3.7190 &3.1645 &3.1608 &3.6915 &3.6355 &3.7149 &3.6672 & 3.5115 & 3.6898 &\sbest{3.7385}& \best{3.8463}\\

\hline
% =====RealLR250=====

\multirow{5}{*}{\begin{tabular}[c]{@{}c@{}}
\textit{\textbf{RealLQ250}}\end{tabular}}

& CLIPIQA $\uparrow$ & 0.7155 &0.5156  &0.6998 &0.6241 &0.6701 &0.6723 &0.7056 &\sbest{0.7369} &0.6899  & 0.5260 & 0.7030 &0.6539& \best{0.7947} \\
& MUSIQ $\uparrow$ & 71.7407 &57.485  &70.161 &61.193 &68.018 &69.553 &71.245 &\sbest{73.217} &69.049  & 62.276 & 71.624 &70.044& \best{73.386} \\
& MANIQA $\uparrow$ & \sbest{0.6301} &0.5116  &0.5807 &0.5027 &0.5080 &0.5782 &0.6053 &0.5924 &0.6031  & 0.4851 & 0.6095 &0.6136& \best{0.6438} \\
& TOPIQ $\uparrow$ & 0.6427 &0.4464   &\sbest{0.6592}  &0.5111 &0.6340 &0.6075 &0.6506 &0.6468 &0.6355  & 0.4907 & 0.6447 &0.5942& \best{0.7299} \\
& QALIGN $\uparrow$ & 4.2144 &3.5499 &4.1325 &3.5917 &3.7199 &4.2486 &4.2061 &4.1772 &4.2066 & 2.8910 & 4.3877 &\sbest{4.2909}& \best{4.3183} \\

\bottomrule
\end{tabular}
}
\label{tab:compare_with_other_methods}
% \vspace{-3mm}
\end{table*}

Spatial feature drifting aligns the distribution of reconstructed features with the HQ feature distribution, thereby improving perceptual quality. However, Real-ISR also requires the restored image to preserve the structural information contained in the LQ input. Unconstrained drifting may introduce visually plausible details that are inconsistent with the input structure. To address this issue, we propose Structure-Modulated Guidance (SMG), which estimates local structural consistency and adaptively balances perceptual enhancement and structural fidelity across spatial locations.

We first characterize local structure through image gradients. For an image $\mathbf{x}$, we convert it to luminance $\mathbf{Y}(\mathbf{x})$ and compute its gradient magnitude:
\begin{equation}
\mathbf{\Gamma}(\mathbf{x})
=
\sqrt{
\left(K_x * \mathbf{Y}(\mathbf{x})\right)^2
+
\left(K_y * \mathbf{Y}(\mathbf{x})\right)^2
},
\end{equation}
where $K_x$ and $K_y$ denote horizontal and vertical Sobel filters, respectively. We then measure the local structural consistency between two spatially aligned images $\mathbf{x_1}$ and $\mathbf{x_2}$ as
\begin{equation}
\mathcal{C}(\mathbf{x_1}, \mathbf{x_2})
=
\frac{
\mathcal{A}
\left[
\mathbf{\Gamma}(\mathbf{x_1}) \odot \mathbf{\Gamma}(\mathbf{x_2})
\right]
}{
\sqrt{
\mathcal{A}
\left[
\mathbf{\Gamma}(\mathbf{x_1})^2
\right]
\odot
\mathcal{A}
\left[
\mathbf{\Gamma}(\mathbf{x_2})^2
\right]
}
},
\end{equation}
where $\mathcal{A}(\cdot)$ denotes local average pooling over a fixed neighborhood, and $\odot$ denotes element-wise multiplication. We then measure the local structural consistency between two spatially aligned images:
\begin{equation}
\mathbf{s}_{\mathrm{pred}}
=
\mathcal{C}(\hat{\mathbf{x}}_{H}, \mathbf{x}_{L}),
\quad
\mathbf{s}_{\mathrm{gt}}
=
\mathcal{C}(\mathbf{x}_{H}, \mathbf{x}_{L}).
\end{equation}
The difference between these two maps captures the local structural deviation of the reconstruction from the HQ reference. We use this difference to define the spatial modulation map:
\begin{equation}
\mathbf{M}
=
\mathrm{clip}
\left(
1
+
\alpha
\left(
\mathbf{s}_{\mathrm{pred}}
-
\mathbf{s}_{\mathrm{gt}}
\right),
m_{\min},
1
\right),
\end{equation}
where $\alpha$ controls the sensitivity to structural deviation, and $m_{\min}$ specifies the lower bound of the modulation coefficient. When the reconstructed structure exhibits similar consistency with the LQ input as the HQ reference, $\mathbf{M}$ remains close to 1. When the reconstructed structure becomes less consistent with the LQ input than the HQ reference, the corresponding modulation coefficient decreases.

Since $G_{\theta}$ starts directly from the LQ input without introducing noise corruption and is trained with reconstruction objectives, the overall content and coarse structure are largely preserved. However, the original drifting field may still introduce high-frequency details that are inconsistent with the input structure. To provide additional structural guidance, we construct a low-frequency drifting field that emphasizes coarse image content while reducing the influence of high-frequency details. Specifically, DWT is applied to $\hat{x}_H$ and $x_H$ to obtain $\hat{x}_H^{\mathrm{low}}$ and $x_H^{\mathrm{low}}$. The same teacher feature extraction and spatial drifting procedure then yields $Q_{\mathrm{low}}^{\ell}$ and $P_{\mathrm{low}}^{\ell}$. The low-frequency drifting field is defined as
\begin{equation}
V_{\mathrm{low}}^{\ell}
=
V_{P_{\mathrm{low}}^{\ell},Q_{\mathrm{low}}^{\ell}}
\left(
Q_{\mathrm{low}}^{\ell}
\right).
\end{equation}

With the original and low-frequency drifting fields, we employ the structure modulation map $\mathbf{M}$ to adaptively balance their contributions across spatial locations. For each selected layer $\ell$, we resize $\mathbf{M}$ to match the spatial resolution of the corresponding teacher features, yielding $\mathbf{m}^{\ell}$. The structure-modulated drifting field is formulated as
\begin{equation}
V_{\mathrm{SMG}}^{\ell}
=
\mathbf{m}^{\ell}
\odot
V^{\ell}
+
\left(
1-\mathbf{m}^{\ell}
\right)
\odot
V_{\mathrm{low}}^{\ell}.
\end{equation}

Regions with larger modulation values receive stronger guidance from the original drifting field for perceptual detail refinement, whereas regions with smaller values rely more on the low-frequency drifting field to suppress structurally inconsistent details. Finally, we incorporate the structure-modulated drifting field into a unified drifting objective and aggregate it over all selected teacher layers:
\begin{equation}
\mathcal{L}_{\mathrm{drift}}
=
\sum_{\ell \in \mathcal{S}}
\frac{1}{N_{\ell}C_{\ell}}
\left\|
Q^{\ell}
-
\operatorname{sg}
\left(
Q^{\ell}
+
V_{\mathrm{SMG}}^{\ell}
\right)
\right\|^{2}.
\end{equation}

This formulation adaptively balances perceptual detail refinement and structural guidance. By modulating the drifting direction before constructing the optimization target, SMG integrates perceptual enhancement and structural guidance within a single drifting objective.

\begin{figure*}[t]
    \centering
    \includegraphics[width=\linewidth]{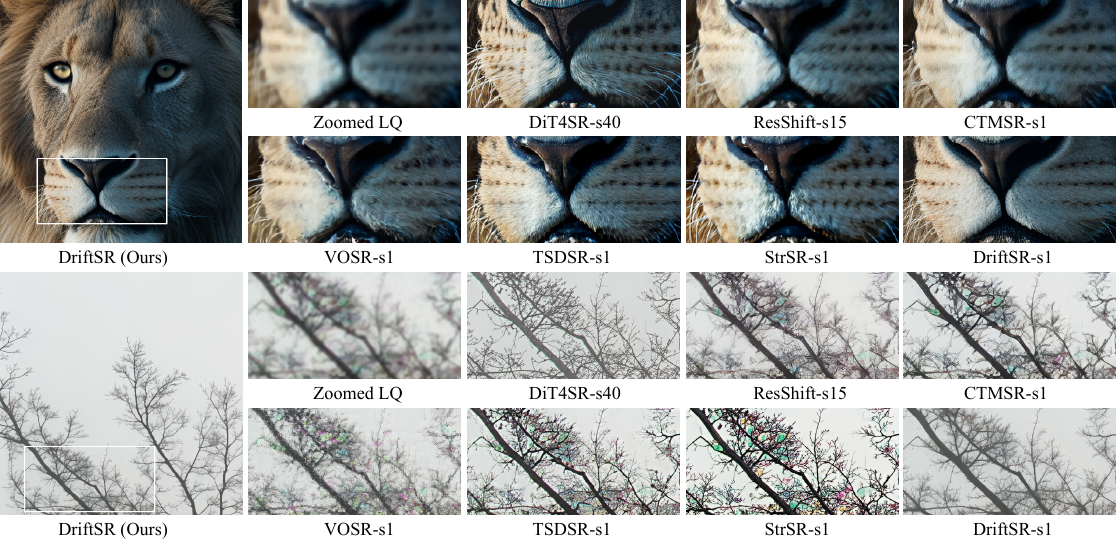}
    % \vspace{-5mm}
    \caption{
        Visual comparison of different methods. $s$ denotes the number of inference steps. Please zoom in for a better view.
    }
    \label{fig:visual comparisons}
    % \vspace{-5mm}
\end{figure*}

\subsection{Training Objective}
\label{sec:training_loss}
During training, only the generator $G_\theta$ is updated and optimized under the joint supervision of the reconstruction loss $\mathcal{L}_{\mathrm{rec}}$ and the drifting loss $\mathcal{L}_{\mathrm{drift}}$:
\begin{equation}
\mathcal{L}
=
\mathcal{L}_{\mathrm{rec}}
+
\lambda_1 \mathcal{L}_{\mathrm{drift}}.
\end{equation}
The data fidelity term $\mathcal{L}_{\mathrm{data}}$ is composed of the MSE loss and LPIPS loss:
\begin{equation}
\mathcal{L}_{\mathrm{rec}}
=
\mathcal{L}_{\mathrm{MSE}}(\hat{x}_H, x_H)
+
\lambda_2
\mathcal{L}_{\mathrm{LPIPS}}(\hat{x}_H, x_H),
\label{eq:rec}
\end{equation}
where $\lambda_1$ and $\lambda_2$ control the contributions of the drifting and perceptual reconstruction objectives, respectively.
\section{Experiments}
\label{sec:experiments}
\subsection{Experimental Settings}
\label{sec:experimental settings}

% ----------------datasets-----------------
\noindent
\textbf{Training and Evaluation Datasets.} We train DriftSR on LSDIR \cite{lsdir} and construct synthetic LR-HR pairs using the Real-ESRGAN degradation pipeline \cite{2021Real}. Degradation-aware prompt extraction (DAPE)~\cite{wu2024seesr} is adopted to generate textual descriptions. For evaluation, we conduct experiments on three real-world benchmarks, including RealSR \cite{2020Toward}, DRealSR \cite{drealsr} and RealLQ250~\cite{2024dreamclear}. Implementation details are provided in the \textbf{supplementary material}.

% ----------------compared methods-----------------
\noindent
\textbf{Compared Methods.} We compare our method with state-of-the-art diffusion-based methods, including the one-step methods OSEDiff~\cite{2024OSEDiff}, CTMSR~\cite{CTMSR}, PiSA-SR~\cite{pisasr}, TSD-SR~\cite{dong2024tsd}, HYPIR~\cite{hypir}, VOSR~\cite{Wu_vosr}, FiDeSR~\cite{kim2026fidesr}, and StrSR~\cite{wang2026strsr}, as well as the multi-step methods StableSR~\cite{2024stablesr}, ResShift~\cite{resshift}, DiT4SR~\cite{dit4sr}, and SeeSR~\cite{wu2024seesr}. Specifically, DiT4SR and TSD-SR are both built on SD3.5-Medium~\cite{sd3}, whereas StrSR adopts FLUX.2-klein-4B~\cite{FLUX.2-klein-4B}. ResShift, CTMSR, and VOSR are developed without T2I priors. 
% All other baselines, as well as our method, are built upon SD2.1.

% All comparative results are obtained using publicly released code and model weights for testing. Comparisons with GAN-based methods are provided in the \textbf{supplementary material}.

% ----------------evaluation metrics-----------------
\noindent
\textbf{Evaluation Metrics.} We comprehensively evaluate all methods using both full-reference and no-reference metrics. Specifically, PSNR and SSIM~\cite{ssim} are used to measure reconstruction fidelity, while LPIPS~\cite{lpips} is used to assess perceptual similarity. For no-reference perceptual quality assessment, we report MUSIQ~\cite{2021MUSIQ}, MANIQA~\cite{2022MANIQA}, CLIPIQA~\cite{clipiqa}, TOPIQ~\cite{2024TOPIQ}, and QALIGN~\cite{QALIGN}. In particular, QALIGN leverages a large multimodal model for image quality assessment and exhibits stronger alignment with human perceptual judgments. We also report the results of a user study in the \textbf{supplementary material}.

% % ----------------implementation details-----------------
% \noindent
% \textbf{Implementation Details.}
% We optimize both the Student Model and the LoRA Model with AdamW~\cite{2019decoupledweight} using a fixed learning rate of 5e-5, and we set the LoRA rank to 4. The models are initialized from SD2.1-base~\cite{sd21-base} and fine-tuned for 6k iterations on four NVIDIA 5880 GPUs with a batch size of 8. Time-Aware VAE is initialized from TADSR~\cite{tadsr}. For the Student Model, the loss weights are set to $\lambda_1=2$, $\lambda_2=1$, $\lambda_3=1$, and $\lambda_4=0.1$, while $\lambda_s$ in \cref{eq:stsd} is set to 0.5. The discriminator is optimized with AdamW using a learning rate of 5e-6, with $\lambda_R=5$ and $\delta=0.005$.

% \textit{Architectural details} are provided in the 

\begin{figure*}[t]
    \centering

    % ================= Left figure =================
    \begin{minipage}[t]{0.49\textwidth}
        \centering

        \includegraphics[width=\linewidth]{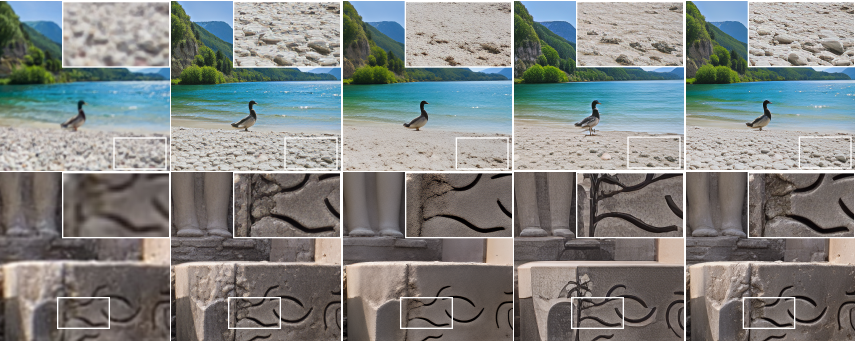}

        % \vspace{-2mm}

        {\scriptsize
\makebox[\linewidth][c]{%
    \makebox[0.2\linewidth][c]{Zoomed LR}%
    \makebox[0.2\linewidth][c]{VGG}%
    \makebox[0.2\linewidth][c]{DINOv2}%
    \makebox[0.2\linewidth][c]{DINOv3}%
    \makebox[0.2\linewidth][c]{DriftSR}%
}}
        % \vspace{-1mm}
        \captionof{figure}{
            Qualitative comparison of different feature representations.
        }
        \label{fig:feature_space}
    \end{minipage}
    % \vspace{-4mm}
    \hfill
    % ================= Right figure =================
    % \vspace{-8mm}
    \begin{minipage}[t]{0.49\textwidth}
        \centering
        \includegraphics[width=\linewidth]{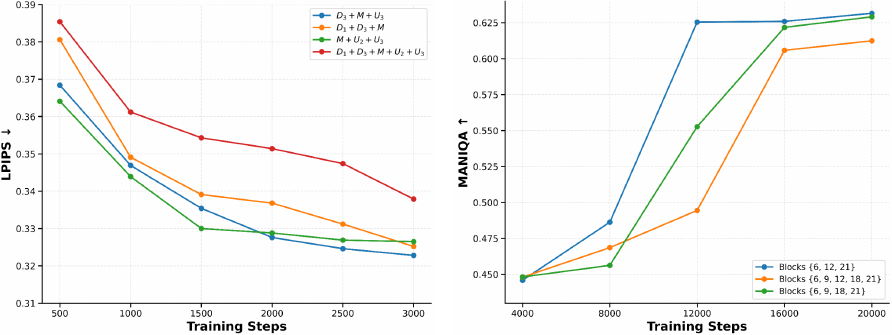}

        {\footnotesize
    \makebox[\linewidth][c]{%
    \makebox[0.6\linewidth][c]{(a) U-Net feature layer configurations.\qquad}%
    \makebox[0.4\linewidth][c]{(b) DiT feature layer configurations.}%
}}

        % \footnotesize (a) different layers in UNet \qquad\qquad\qquad (b) different layers in DiT
        \vspace{-0.5mm}
        \captionof{figure}{
            Ablation study on teacher feature layer selection.
        }
        \label{fig:teacher_layer_ablation}
    \end{minipage}

\end{figure*}
\begin{table*}
\centering
\scriptsize
\caption{Comparison of inference steps, runtime, and model parameters among different Real-ISR methods on $512\times512$ inputs.}
\label{tab:complexity}
\vspace{-2mm}
\fontsize{7pt}{8.4pt}\selectfont
\setlength{\tabcolsep}{3.2pt}
\renewcommand{\arraystretch}{1.1}
\begin{tabular*}{\textwidth}{@{\extracolsep{\fill}}l
    ccc
    cccc
    ccc
    cc@{}}
\toprule
 & \multicolumn{4}{c}{\textit{Multi-step Real-ISR Methods}} 
 % & \multicolumn{4}{c}{\textit{Multi-step $\cdot$ T2I-based}}
 % & \multicolumn{3}{c}{\textit{One-step $\cdot$ Vision-only}}
 & \multicolumn{8}{c}{\textit{One-step Real-ISR Methods}}\\
\cmidrule(lr){2-5}\cmidrule(lr){6-13}
 & StableSR & DiT4SR & SeeSR
 & ResShift & CTMSR & OSEDiff & PiSA-SR
 & TSDSR & HYPIR & VOSR-1.4B
 & FiDeSR & DriftSR \\
\midrule
Inference Step
 &  200 & 40 & 50
 & 15 & 1 & 1 & 1
 & 1  & 1  & 1
 & 1 & 1 \\
Params (M)
 &  1{,}563  &  2{,}716 & 2{,}514
 & 119 & 171 & 1{,}775 & 1{,}302
 &  2{,}207  & 1{,}549  & 1{,}745
 & 1{,}291 & 1{,}777 \\
Time (s)
 &  14.154  & 6.947  & 6.365
 &  1.471 & 1.312 &  0.191 & 0.212
 &  0.122  & 0.275  & 0.174
 &  0.171  &  0.195 \\
\bottomrule
\end{tabular*}
\vspace{-2mm}
\end{table*}

% Params (M)
%  & 174.0   & 619.4  & 1{,}742.9
%  & 1{,}540 & 1{,}680 & 2{,}510 & 21{,}380
%  & 174.0   & 620.7  & 1{,}745.7
%  & 1{,}760 & 1{,}290 \\
% Time (s)
%  & 0.534   & 1.226  & 2.114
%  & 10.036  & 2.751  & 3.943  & 12.550
%  & 0.116   & 0.094  & 0.095
%  & 0.104   & 0.091  \\

% ----------Comparison with State-of-the-Arts
\subsection{Comparison with State-of-the-Arts}
\label{comparison}

\noindent
\textbf{Quantitative Comparisons.} Table~\ref{tab:compare_with_other_methods} compares DriftSR with existing diffusion-based methods on three real-world datasets. The results yield three main observations. (1) DriftSR achieves the best overall performance on no-reference perceptual metrics across all three benchmarks. The advantage is particularly pronounced on the higher-resolution RealLQ250 dataset, indicating that DriftSR can more effectively leverage the rich generative priors of pretrained diffusion models to recover realistic perceptual details. (2) ResShift, CTMSR, and VOSR employ purely visual architectures without T2I diffusion priors. Although these methods achieve competitive performance on full-reference metrics such as PSNR and SSIM, they generally underperform DriftSR on no-reference perceptual metrics, suggesting limitations in recovering fine-grained high-frequency details. (3) Despite using the same SD2.1 prior, FiDeSR favors fidelity-oriented reconstruction, whereas DriftSR achieves substantially better perceptual performance, particularly on CLIPIQA and TOPIQ.

Meanwhile, DriftSR does not consistently achieve the best scores on full-reference metrics, while maintaining strong perceptual performance. This observation is consistent with the recognized trade-off between fidelity and perceptual quality~\cite{2018perception, 2019wavelet}. As richer perceptual details are recovered, full-reference metrics may not fully reflect perceptual quality and often favor smoother reconstructions~\cite{supir, wu2024seesr}.

% This difference reflects the established fidelity-perception trade-off: full-reference metrics may favor smoother reconstructions and diverge from human perception as richer details emerge~\cite{wu2024seesr, dong2024tsd, 2018perception}.

% ----------------qualitative comparisons----------------
\noindent
\textbf{Qualitative Comparisons.} Figure~\ref{fig:visual comparisons} presents visual comparisons among different methods. In the first row, DriftSR reconstructs more natural and perceptually realistic textures, particularly for fine animal fur. Although ResShift, CTMSR, and VOSR achieve competitive PSNR and SSIM scores, their results exhibit noticeable over-smoothing. In the second row, DriftSR effectively suppresses degradation artifacts and recovers clearer structures with consistent colors, whereas other one-step methods show limited recovery of fine structures and fail to correct color distortions. These comparisons further demonstrate the stronger capability of DriftSR to leverage the generative prior. More visual comparisons are provided in the \textbf{supplementary material}.

% ----------------complexity comparisons-----------------
\noindent
\textbf{Complexity Comparisons.}
We further evaluate the computational efficiency of representative Real-ISR methods in Table~\ref{tab:complexity}, considering inference steps, runtime, and model parameters. We measure inference time on the $\times4$ SR task with $128\times128$ LQ inputs, applying bicubic upsampling before inference on a single NVIDIA 5880 GPU. DriftSR substantially reduces inference cost compared with multi-step methods. It is more than 72 times faster than StableSR, 35 times faster than DiT4SR, over 32 times faster than SeeSR, and 7 times faster than ResShift. Compared with existing one-step models, DriftSR matches OSEDiff in inference speed and is 6 times faster than CTMSR. Despite being slower than VOSR, TSD-SR, and FiDeSR, DriftSR achieves a favorable balance between computational efficiency and perceptual restoration quality.

% ----------------quantitative comparisons-----------------

% ----------------Ablation Study-----------------
\subsection{Ablation Study}
\label{sec:ablation_study}

% ------------- loss ablation ---------------
\noindent
\textbf{Effectiveness of Drifting and SMG.} We conduct an ablation study to evaluate the contribution of drifting and Structure-Modulated Guidance (SMG), as shown in Table~\ref{tab:abla_loss}. Introducing drifting consistently improves all no-reference perceptual metrics on both RealSR and DRealSR, demonstrating its effectiveness in enhancing perceptual quality. Compared with the variant without SMG, the full DriftSR achieves lower LPIPS on both datasets, showing that SMG effectively regulates the drifting process and improves perceptual similarity.

\begin{figure*}[t]
    \centering
    \includegraphics[width=\linewidth]{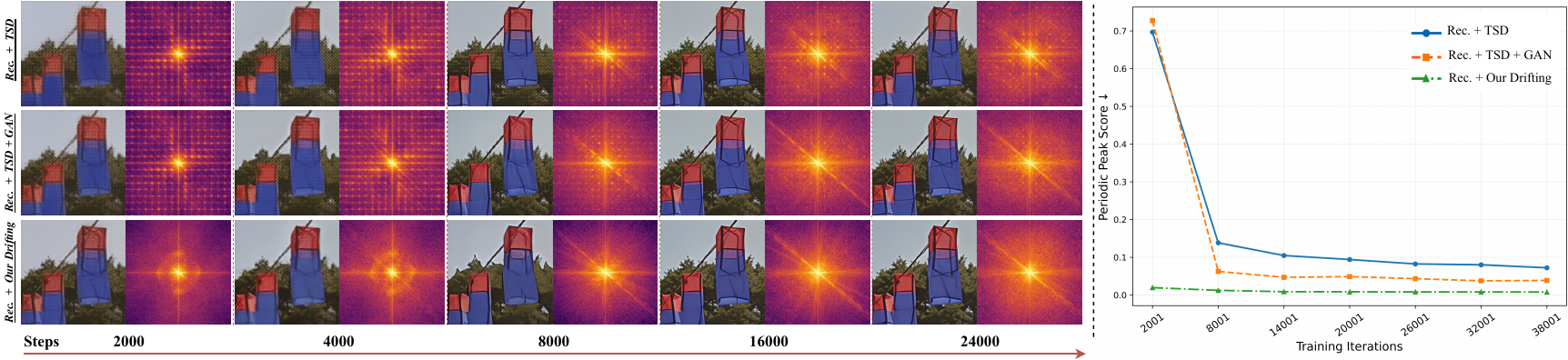}
    {\footnotesize
    \makebox[\linewidth][c]{%
    \makebox[0.7\linewidth][c]{(a) Qualitative results and corresponding 2D Fourier spectra over training steps.}%
    \makebox[0.3\linewidth][c]{(b) Periodic Peak Score over training steps.}%
}}
    \vspace{-4mm}
    \caption{Periodic artifact suppression with the pretrained SD3.5-Medium prior. (a) Comparison across training steps using the reconstruction objective with TSD, additional GAN supervision, and our Drifting loss. (b) Periodic Peak Score across training steps, computed following OP4KSR~\cite{op4ksr}, where lower scores indicate fewer periodic artifacts.}
    \label{fig:train_visual}
    \vspace{-2mm}
\end{figure*}
\begin{table*}[t]
% \vspace{-0.2cm}
  \centering
  % ===== Left table =====
\begin{minipage}[t]{0.5\linewidth}
  \centering
  \caption{\small{Ablation study of Spatial Feature Drifting (SFD) and Structure-Modulated Guidance (SMG) on RealSR and DRealSR.}}
  \label{tab:abla_loss}
  \begingroup
  \resizebox{1\linewidth}{!}{%
    \begin{tabular}{c|c|ccccc}
    \toprule
    Dataset & Method & LPIPS$\downarrow$ & CLIPIQA$\uparrow$ & MUSIQ$\uparrow$ & MANIQA$\uparrow$ & TOPIQ$\uparrow$ \\
    \midrule
    \multirow{4}{*}{RealSR} 
    & w/o SFD  & 0.3189  & 0.6816 & 68.64 & 0.6284 & 0.6739 \\
    & w/o SMG  & 0.3395 & 0.7407 & 70.23 & 0.6521 & 0.7127 \\
    % & w/o Neg  & 0.3358 & 0.7313 & 70.45 & 0.6626 & 0.7213 \\
    & \cellcolor{gray!20}DriftSR  & \cellcolor{gray!20}0.3287 & \cellcolor{gray!20}0.7334 & \cellcolor{gray!20}71.28 & \cellcolor{gray!20}0.6659 & \cellcolor{gray!20}0.7275 \\
    \midrule 
    \multirow{4}{*}{DRealSR} 
    & w/o SFD & 0.3247 & 0.7015 & 64.13 & 0.5842 & 0.6502 \\
    & w/o SMG  & 0.3493 & 0.7538 & 68.07 & 0.6160 & 0.6852\\
    % & w/o Neg  & 0.3451 & 0.7437 & 67.71 & 0.6234 & 0.7004 \\
    & \cellcolor{gray!20}DriftSR  &\cellcolor{gray!20}0.3379 & \cellcolor{gray!20}0.7442 & \cellcolor{gray!20}67.49 & \cellcolor{gray!20}0.6226 & \cellcolor{gray!20}0.6970 \\
    \bottomrule
  \end{tabular}
  }
  \endgroup
\end{minipage}
  \hfill
  \begin{minipage}[t]{0.47\linewidth}
    \centering
    \vspace{-1.5mm}
    \caption{\small{Ablation study on different feature spaces.}}
    \label{tab:feature_space}
    \resizebox{1\linewidth}{!}{%
      \begin{tabular}{c|c|ccccc}
    \toprule
    Dataset & Space & LPIPS$\downarrow$ & CLIPIQA$\uparrow$ & MUSIQ$\uparrow$ & MANIQA$\uparrow$ & TOPIQ$\uparrow$ \\
    \midrule
    \multirow{4}{*}{RealSR} 
    & VGG & 0.3312 & 0.7005 & 69.42 & 0.6420 & 0.7018 \\
    % & MAE  & & & & & \\
    & DINOv2 & 0.3213 & 0.6993 & 69.92 & 0.6588 & 0.7086 \\
    & DINOv3 & 0.4171 & 0.7380 & 71.17 & 0.6584 & 0.7420 \\
    & \cellcolor{gray!20}Teacher  & \cellcolor{gray!20}0.3287 & \cellcolor{gray!20}0.7334 & \cellcolor{gray!20}71.28 & \cellcolor{gray!20}0.6659 & \cellcolor{gray!20}0.7275 \\
    \midrule 
    \multirow{4}{*}{DRealSR} 
    & VGG  & 0.3343 & 0.7110 & 65.53 & 0.5955 & 0.6756 \\
    % & MAE  & & & & & \\
    & DINOv2  & 0.3294 & 0.7199 & 66.64 & 0.6173 & 0.6900 \\
    & DINOv3  & 0.4222 & 0.7373 & 68.41 & 0.6332 & 0.7149 \\
    & \cellcolor{gray!20}Teacher &\cellcolor{gray!20}0.3379 & \cellcolor{gray!20}0.7442 & \cellcolor{gray!20}67.49 & \cellcolor{gray!20}0.6226 & \cellcolor{gray!20}0.6970 \\
    \bottomrule
  \end{tabular}
    }
\end{minipage}
\vspace{-0.4cm}
\end{table*}

\noindent
\textbf{Effect of Feature Space.} We ablate the feature space used to compute the drifting loss by replacing the teacher representation with VGG~\cite{vgg}, DINOv2~\cite{dinov2}, and DINOv3-ConvNeXt~\cite{dinov3}, with three layers selected for each variant. As shown in Table~\ref{tab:feature_space}, DINOv3-ConvNeXt provides strong perceptual representations but yields worse LPIPS scores of 0.4171 and 0.4222 on RealSR and DRealSR, respectively. Visual comparisons in Figure~\ref{fig:feature_space} also show larger structural deviations from the LQ input, suggesting that stronger semantic representations are not necessarily well aligned with the structural constraints of image restoration. In contrast, the teacher feature space consistently outperforms VGG and DINOv2 on perceptual metrics across both benchmarks, indicating that it provides a more suitable representation space for drifting and a better balance between perceptual quality and structural fidelity.

% ------------- feature layers ---------------
\noindent
\textbf{Effect of Teacher Layer Configurations.} We further ablate the teacher layers used to compute the Drifting loss in the U-Net of SD2.1-Base and the MMDiT of SD3.5-Medium, as shown in Figure~\ref{fig:teacher_layer_ablation}. We report LPIPS for the U-Net and MANIQA for the MMDiT, as these metrics exhibit clearer trends across layer configurations. For the U-Net, features from different stages operate at different spatial scales and encode distinct representations. As shown in Figure~\ref{fig:teacher_layer_ablation} (a), \(D_i\) and \(U_i\) denote the \texttt{downsampling} and \texttt{upsampling} blocks indexed by \(i\), respectively, while \(M\) denotes the \texttt{bottleneck} block. Different combinations of three layers yield only minor performance variations. In contrast, the number of selected layers has a larger impact, as increasing the number from three to five degrades the performance of DriftSR. For the MMDiT backbone, all transformer blocks operate at the same spatial resolution and differ mainly in network depth. Prior work~\cite{dit_block_14} has shown that intermediate DiT blocks can exhibit stronger representations of class semantics. We therefore further investigate layer selection in DriftSR. As shown in Figure~\ref{fig:teacher_layer_ablation} (b), including the intermediate \texttt{Block 12} provides no performance benefit. Instead, the results suggest that performance is more sensitive to the number of selected layers.

% ------------- suppresses checkerboard artifacts ---------------
\noindent
\textbf{Drifting Loss Suppresses Checkerboard Artifacts.} With the pretrained SD3.5-Medium prior, the patch tokenization in MMDiT may contribute to periodic checkerboard artifacts. As shown in Figure~\ref{fig:train_visual} (a), we visualize the restoration results and corresponding 2D Fourier spectra at different training steps. Combining reconstruction loss (Rec.) with TSD~\cite{dong2024tsd} produces persistent periodic structures in the frequency domain. Adding GAN~\cite{omgsr} supervision gradually reduces these artifacts during training but does not prevent their emergence at early stages. In contrast, replacing TSD with our Drifting loss effectively suppresses dense periodic patterns throughout training. Using the autocorrelation formulation in OP4KSR~\cite{op4ksr} as a reference, we further compute the Periodic Peak Score, as shown in Figure~\ref{fig:train_visual} (b), which quantitatively confirms the suppression of periodic artifacts by the drifting loss. Details of the Periodic Peak Score computation are provided in the \textbf{supplementary material}.

% 基于MMDiTi架构的图像恢复任务在训练过程中容易产生周期性的棋盘伪影。如图6（a）所展示的训练过程中的恢复结果和2D傅里叶频谱,在使用 SD3.5-Medium 预训练先验时，重建损失（Rec.）与 TSD 的组合在训练过程中会产生持续的周期性结构，加入 GAN 监督后，这些周期伪影随着训练推进而有所减弱，但并不能阻止其在训练早期形成。相比之下，相比之下，将 TSD 替换为我们的 Drifting loss 后，整个训练过程中频谱中的密集周期性网格均得到明显抑制。进一步地，我们参考 OP4KSR 中基于自相关的周期性分析方式计算右侧的 Periodic Peak Score，在定量上进一步验证了 Drifting loss 对周期伪影的持续抑制作用。

% \section{Conclusion and Limitation}
\section{Conclusion}
\label{sec:conclusion}
In this paper, we presented DriftSR, a one-step Real-ISR framework that exploits pretrained diffusion priors through distribution drifting, with only the one-step generator optimized during training. We introduced Spatial Feature Drifting, which reformulates distribution drifting at the spatial feature level by treating each spatial location as a distributional sample. We further introduced Structure-Modulated Guidance, which adaptively refines the drifting guidance based on local structural consistency with the LQ input. Extensive experiments on three real-world benchmarks demonstrate that DriftSR achieves strong perceptual quality with efficient one-step inference, demonstrating the effectiveness of distribution drifting for Real-ISR.

% \noindent
% \textbf{Limitations.} We need to train a teacher model to provide boundary supervision, which inevitably increases the training cost and compute requirements. In future work, we plan to develop teacher-free and more efficient distillation strategies to further reduce training overhead while maintaining perceptual quality.

{
    \small
    % \FloatBarrier
    \bibliographystyle{ieeenat_fullname}
    \bibliography{main}
}

% \input{sec/supp}

\end{document}